\documentclass[lettersize,journal]{IEEEtran}

\usepackage{amsmath,amssymb,amsopn,amstext,amsfonts}

\usepackage{array}      
\usepackage{tabularx}   
\usepackage{booktabs}
\usepackage{multirow}
\usepackage{multicol}

\usepackage{graphicx}
\usepackage{stfloats}
\usepackage[caption=false,font=normalsize,labelfont=sf,textfont=sf]{subfig}

\usepackage{algpseudocode}  
\usepackage{algorithm}

\usepackage{textcomp}
\usepackage{url}
\usepackage{verbatim}
\usepackage{cite}
\usepackage{pifont}    
\usepackage{utfsym}

\usepackage[linkcolor=blue,citecolor=blue,urlcolor=blue,colorlinks=true]{hyperref}
\usepackage{cleveref}

\usepackage{tablefootnote}

\graphicspath{{./Figures/}}
\DeclareGraphicsExtensions{.pdf,.png,.jpg,.eps,.svg}
\IEEEoverridecommandlockouts

\title{IRS: Instance-Level 3D Scene Graphs via Room Prior Guided LiDAR-Camera Fusion
}
\author{
Hongming Chen$^*$, Yiyang Lin$^*$, Ziliang Li, Biyu Ye, Yuying Zhang, Ximin Lyu
    \thanks{* means equal contribution.}
    \thanks{Corresponding author: Ximin Lyu}
    \thanks{All authors are with the School of Intelligent Systems Engineering, Sun Yat-sen University, Guangzhou, China. }
        \thanks{E-mail:{\tt\small chenhm223@mail2.sysu.edu.cn }}
        
}
\begin{document}
\maketitle

\begin{abstract}
    Indoor scene understanding remains a fundamental challenge in robotics, with direct implications for downstream tasks such as navigation and manipulation. Traditional approaches often rely on closed-set recognition or loop closure, limiting their adaptability in open-world environments. With the advent of visual foundation models (VFMs), open-vocabulary recognition and natural language querying have become feasible, unlocking new possibilities for 3D scene graph construction.
    In this paper, we propose a robust and efficient framework for instance-level 3D scene graph construction via LiDAR-camera fusion. Leveraging LiDAR's wide field of view (FOV) and long-range sensing capabilities, we rapidly acquire room-level geometric priors. Multi-level VFMs are employed to improve the accuracy and consistency of semantic extraction. During instance fusion, room-based segmentation enables parallel processing, while the integration of geometric and semantic cues significantly enhances fusion accuracy and robustness. Compared to state-of-the-art methods, our approach achieves up to an order-of-magnitude improvement in construction speed while maintaining high semantic precision.
    Extensive experiments in both simulated and real-world environments validate the effectiveness of our approach. We further demonstrate its practical value through a language-guided semantic navigation task, highlighting its potential for real-world robotic applications.
\end{abstract}

\section{Introduction}

Semantic navigation is a crucial step toward enabling robots to seamlessly integrate into human living environments. Expectations for robotic systems have evolved beyond simple point-to-point navigation and obstacle avoidance to encompass human-like environmental perception and semantic understanding~\cite{deng2025whole,han2023efficient,gervet2023navigating,wu2021scenegraphfusion}. For instance, we expect to interact with robots through natural language commands such as “go to the wooden chair in the living room” or “find the television in the laboratory,” rather than specifying abstract 3D coordinates.

\begin{figure}[ht]
    \centering
    \includegraphics[width=1\linewidth]{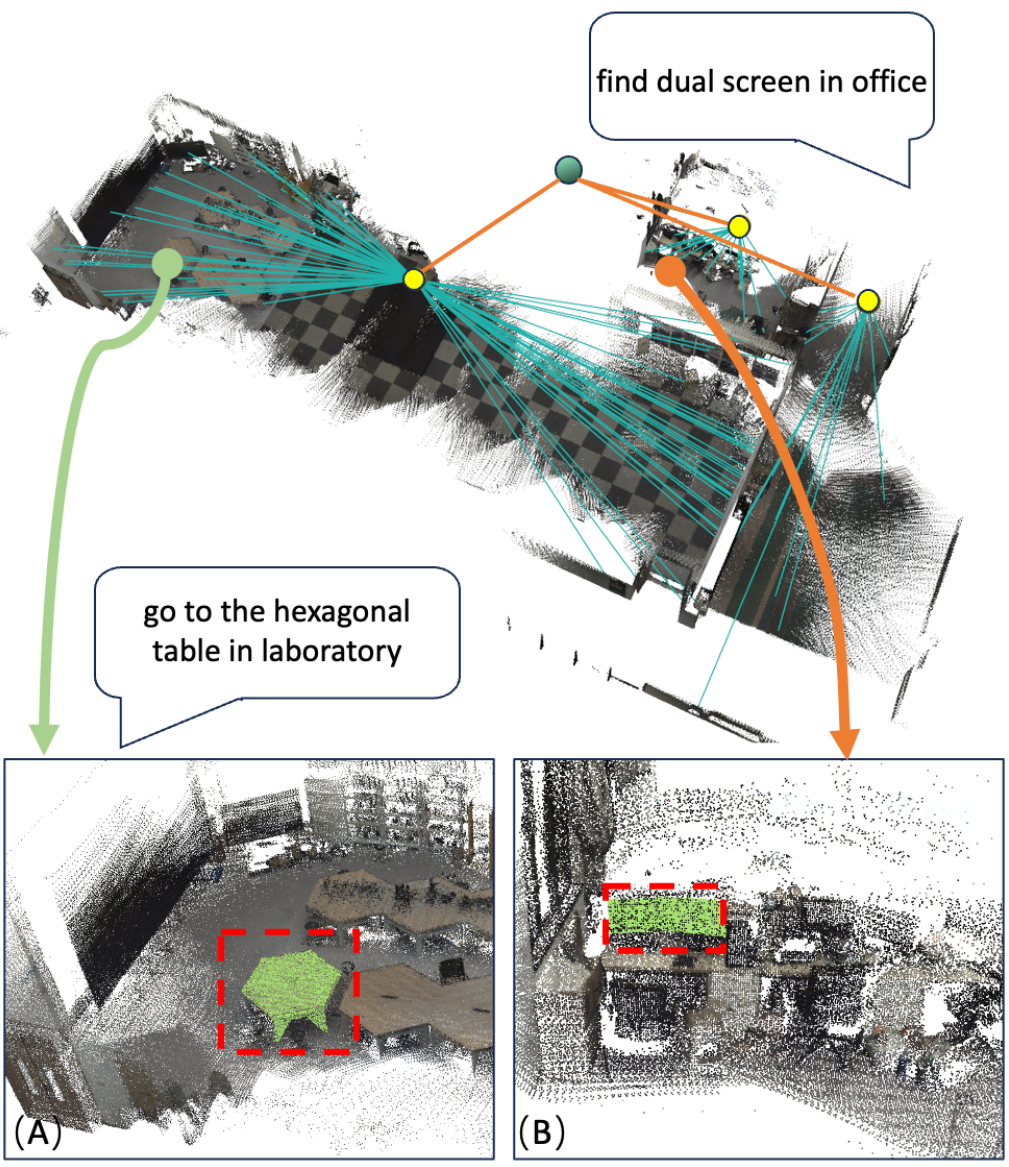}
    \caption{Real-world 3D scene graph query results. We use a large language model (LLM) to query the constructed graph with natural language. (A) shows the response to the prompt \textit{“go to the hexagonal table in the laboratory”}. (B) shows the result for \textit{“find the dual screen in the office”}.}
    \label{fig:1}
    \vspace{-10pt}
\end{figure}

Nevertheless, the realization of these capabilities remains highly challenging. Robots must achieve precise object localization, understand semantic attributes, and interpret human language. This process typically involves multiple stages, including geometric localization, semantic recognition, and cross-modal alignment. In contrast to traditional geometric approaches that rely on ray-casting for robustness, semantic mapping requires consistent multi-view observations and robustness to variations in illumination, sensor noise, and occlusions—significantly increasing the complexity of semantic extraction and fusion~\cite{greve2024collaborative,yang2023sam3dsegment3dscenes}.
Furthermore, existing scene graph construction methods often suffer from prohibitively high computational costs. Constructing a comprehensive 3D scene graph for a single indoor environment can take several hours.

Recently, significant progress has been achieved in this domain, driven by several key developments. These include breakthroughs in SLAM techniques~\cite{rosinol2020kimera, mccormac2018fusion++}, improvements in computer vision systems~\cite{mccormac2017semanticfusion}, and the emergence of VFMs~\cite{kirillov2023segany, liu2023grounding}.
Despite these advances, most existing approaches remain constrained in two critical ways: they rely heavily on closed-set semantic extraction and depend on loop closure detection for robust mapping~\cite{rosinol2020kimera,mccormac2018fusion++}. Although some methods leverage VFMs to build open-vocabulary scene graphs, their performance in real-world scenarios remains suboptimal~\cite{werby2024hierarchical}. Moreover, previous approaches~\cite{rosinol2021kimera} primarily employ depth cameras, which are more vulnerable to noise and illumination conditions than LiDAR.
Our work aims to develop a robust and efficient scene graph construction algorithm capable of operating effectively in real environments, thus laying a solid foundation for downstream tasks such as navigation and manipulation.

In this paper, we present \textbf{IRS} — Instance-Level 3D Scene Graphs via Room Prior Guided LiDAR-Camera Fusion. Our approach integrates RGB images and LiDAR point clouds to build a hierarchical 3D scene graph. Leveraging LiDAR’s 360-degree FOV and long-range sensing capabilities, we first perform room-level geometric segmentation to obtain prior spatial information about object affiliations. Subsequently, multi-level VFMs are utilized both to enhance the accuracy and consistency of mask segmentation and to enable semantic alignment with natural language. During the fusion stage,  room-based segmentation enables parallel processing, while the integration of geometric and semantic cues significantly enhances fusion accuracy and robustness. Ultimately, our method achieves an order-of-magnitude speedup over state-of-the-art approaches in constructing 3D scene graphs.

Our main contributions are as follows:
\begin{itemize}
    \item We introduce a novel framework for 3d scene graph construction that achieves seamless integration of LiDAR and camera modalities, facilitating incremental and open-vocabulary instance-level semantic mapping.
    \item By exploiting the comprehensive 360° FOV provided by LiDAR sensors, we derive room-level structural priors that significantly accelerate the instance-level semantic fusion process, an order-of-
    magnitude computational efficiency compared to current state-of-the-art approaches, while maintaining comparable segmentation fidelity.
    \item We propose a robust fusion strategy that jointly leverages semantic and geometric cues for instance association, thereby significantly enhancing both the precision and efficiency of the fusion stage.

\end{itemize}

\section{Related work}
\subsection{3D Semantic Mapping}
Semantic mapping plays a critical role in enabling robots to interpret and interact with their environments beyond geometric reasoning. It can be represented in various forms. Early approaches often utilize 2D grid maps augmented with additional channels to encode semantic labels~\cite{chang2023goatthing,chaplot2020learningexploreusingactive}. Zhang \textit{et al.}~\cite{lu2024semantics} extend this idea to 3D by employing voxel-based representations in an egocentric mapping framework. While 3D grid maps improve spatial coverage~\cite{whelan2016elasticfusion, rosinol2020kimera, liu2024fm}, they suffer from limited resolution due to memory constraints and rely on closed-set semantic classification~\cite{mccormac2017semanticfusion}, which restricts their flexibility and generalization. 

In contrast, scene graph-based representations offer a multi-level and expressive alternative. Nodes represent semantic objects, and edges capture spatial or semantic relations~\cite{yang2018graph, wu2021scenegraphfusion}. 
More importantly, scene graphs naturally support semantics by associating each object with rich, instance-level features aligned with vision-language models such as CLIP~\cite{radford2021learningtransferablevisualmodels}, enabling open-set querying and language-based interaction~\cite{yokoyama2024vlfm, yin2024sgnavonline3dscene}.
Earlier scene graph methods were constrained to 2D representations, falling short of modeling the complexity of real-world 3D environments~\cite{yang2018graph}. Armeni \textit{et al.}~\cite{armeni20193d} introduced one of the first approaches to parse a metric-semantic 3D mesh into a 3D scene graph. Subsequent works, such as Fusion++~\cite{mccormac2018fusion++} and SlideSLAM~\cite{liu2024slideslam}, incorporated instance-level semantics into 3D graph construction. However, many of these methods rely on coarse, closed-set labels (\textit{e.g.}, “class 1”, “class 2”), which hinder their generalizability—especially in zero-shot settings~\cite{hughes2022hydra}.

With the emergence of foundation models, open-vocabulary understanding has become a growing trend in 3D scene representation~\cite{werby2024hierarchical}. However, some of these methods perform instance merging based on per-mask comparisons, which become increasingly inefficient as the number of instances grows. In contrast, Yang \textit{et al.}~\cite{yang2023sam3dsegment3dscenes} propose a frame-level merging strategy, where runtime remains constant regardless of the number of existing instances. However, these methods do not align object features with vision-language embeddings, and therefore do not support querying through VLMs.

In our work, we use a three-level scene graph: building, room, instance. Each object is represented by its point cloud, preserving fine-grained geometry and supporting instance-aware semantics. This structure enables incremental mapping in large-scale environments and supports natural language querying—key capabilities for interactive robotic systems.

\subsection{Room Segmentation and Layout Estimation}

Dividing indoor spaces into individual rooms or functional units is a key step for many downstream robotic tasks, such as semantic mapping, navigation, and instance association~\cite{bormann2016room}. In our work, room segmentation serves not only as a spatial abstraction but also as a heuristic method to significantly accelerate instance-level semantic fusion.

Traditional room segmentation methods mainly rely on global point clouds and geometric feature extraction, where room division is achieved by analyzing the geometric shapes of different regions in the environment, such as the positions of walls, doors, and windows~\cite{luperto2019predicting, luperto2022robust, he2021hierarchical}. However, these methods face significant limitations in practical applications, particularly in dynamic and complex environments, where room segmentation tends to be coarse, lacking flexibility and adaptability.

\begin{figure*}[ht]
    \centering
    \includegraphics[width=1\linewidth]{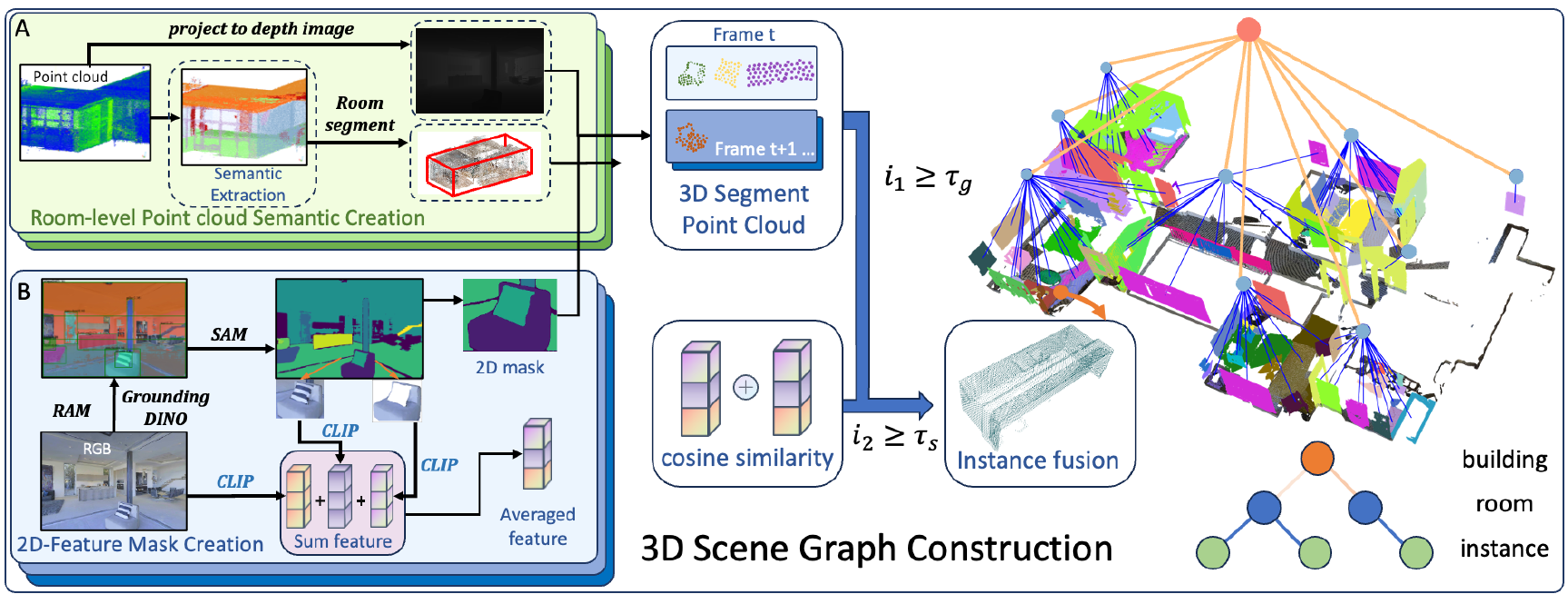}
    \caption{Flowchart of our scene graph construction pipeline. Subfigure A illustrates the preprocessing of the input point cloud to extract room-level geometric information. Subfigure B presents the instance-level semantic extraction process. The right side of the figure shows how to use the obtained room information and semantic information to build a 3D scene graph. } 
    \label{fig:2}
\end{figure*}

Ambrus et al.\cite{ambrucs2017automatic} propose a room-level segmentation method based on multi-class energy minimization over static 3D point clouds, but it lacks online or incremental processing capabilities. To address incremental segmentation, Blochliger \textit{et al.}\cite{blochliger2018topomap} employ Generalized Voronoi Diagrams (GVDs) with boundary dilation. While their method improves segmentation efficiency, it suffers from low spatial resolution and high memory overhead, limiting its applicability in complex or large-scale environments~\cite{chen2022fast}.

Directly extracting semantics from point clouds is an intuitive approach~\cite{qi2017pointnet}, but most point cloud semantic extraction methods require global point clouds (i.e., point clouds from multiple perspectives), which is not feasible in partial observation. Another challenge is that point cloud semantic extraction needs to achieve instance-level classification~\cite{zhang2019review}, rather than simple class classification, or it cannot be directly applied to room segmentation tasks.

In this paper, we build on the OneFormer3D~\cite{kolodiazhnyi2024oneformer3d} method, inputting incrementally constructed point cloud maps to obtain instance-level semantics. Combined with geometric methods, we further achieve precise room segmentation.

\section{Methodology}
We aim to develop a concise and efficient scene graph construction system that generates instance-level, open-vocabulary 3D representations from RGB, point cloud, and odometry inputs—enabling direct interaction via natural language queries.
Our pipeline begins with a point cloud-based room segmentation method that assigns object instances to corresponding spatial regions (Sec.\ref{SECTION III-A}). We then introduce a robust instance-level semantic extraction approach that improves over direct SAM-based segmentation by enhancing resilience to occlusion and illumination changes (Sec.\ref{SECTION III-B}).
To accelerate instance fusion, we exploit room-level geometric priors to constrain and parallelize the matching space. We further integrate semantic and geometric cues during merging to improve fusion accuracy and consistency (Sec.~\ref{SECTION III-C}).
Finally, we build a hierarchical scene graph combining spatial and semantic information (Sec.\ref{SECTION III-D}) and demonstrate its utility through language-guided semantic navigation on a real robot platform. The full framework is illustrated in Fig.~\ref{fig:2}.

\subsection{Room-level Semantic Segmentation}
\label{SECTION III-A}
Rooms serve as vital intermediaries between building and object levels in scene graphs. Unlike bottom-up methods that overlook room geometry, we leverage LiDAR’s 360° FOV to infer room-level semantics, inspired by human spatial reasoning~\cite{tversky2003structures}, enabling faster and more structured scene graph construction.

\begin{algorithm}
\caption{Room Segment}
\begin{algorithmic}[1] 
    \State \textbf{Definition:} Let $O_i$ denote the robot’s position, $P_{O_i}$ the point cloud set to be segmented around $O_i$, $N$ the side length of the segmentation block, and $\mathcal{R}$ the set of registered rooms, $S_j$ is the semantic label, $\mathcal{S}$ is the set of $S_j$, $B_i$ is the bounding box size of room $r_i$.
    
    \State $P_{O_i}\gets \text{kdtree\_cute}(N,O_i),\mathcal{R} \gets \emptyset$
    
    \State $\text{Init\_Room}(P_{O_i})$
    \While{$P_{O_i} \neq \emptyset$}
        \State $\mathcal{S}_O = \{S_0,\dots, S_j\} \gets \text{Semantic\_Extract}(P_{O_i})$
        \State $\mathcal{S} = \text{Assign\_to\_Wall}(\mathcal{S}_{O})$
        \For{$S_j,S_{j+1} \in \mathcal{S}$}
            \State $c_1 = overlap(S_{j-1},S_{j})$
            \State $c_2 = cos(V_{j-1},V_{j})$
            \If{$(c_1 > \tau_1) \land (c_2 \ge \tau_2)$}
                \State \text{Add in the room}
                \State \text{Update $r_i$ size}
            \EndIf
            
        \EndFor
        \If{$\text{No merge in the for loop}$}
            \State $i\mathrel{+{+}},~\mathcal{R} \gets \text{create a new room}:r_i$
        \EndIf
        
    \EndWhile
    
\end{algorithmic}
\label{Algorithm:1}
\end{algorithm}

We adopt the OneFormer3D network~\cite{kolodiazhnyi2024oneformer3d} to extract instance-level semantics from point cloud data. Specifically, we first construct a KD-tree from the accumulated point cloud and then extract a rectangular region of size $N$, centered on the robot in the ego-centric frame. This local region is fed into the network, which outputs semantic labels for all points within it.
Since the collected point cloud data represents only a partial observation of the environment, we retain only the semantic predictions relevant to room structure—namely, \texttt{walls}, \texttt{windows}, \texttt{doors}, \texttt{ceilings}, and \texttt{floors}. Semantic labels associated with other object classes are discarded, as they are less reliable under partial views and not essential for room-level reconstruction. Furthermore, as the network outputs may contain noise and inconsistencies, we adopt a geometry-based approach to merge wall segments. This room-oriented geometric reasoning forms a core contribution of our work.

\begin{figure}[ht]
    \centering
    \includegraphics[width=1\linewidth]{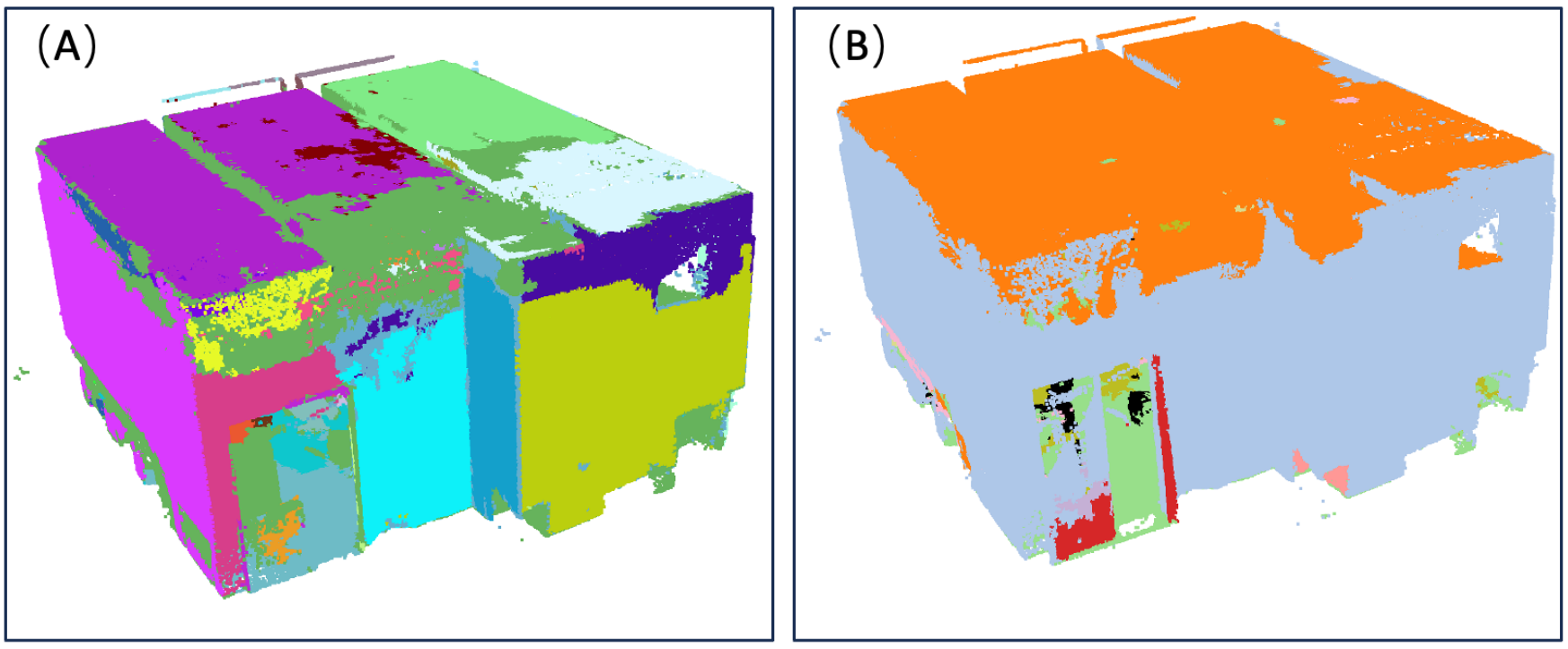}
    \caption{Room segmentation results. (A) shows the semantic segmentation output of the point cloud. (B) presents the final extracted rooms after geometric merging.}
    \label{fig:room segment}
\end{figure}

For the accuracy of the explanation, our pseudo code can be found Alg~\ref{Algorithm:1}. The result can be seen at Fig.~\ref{fig:room segment}. We use overlap detection and feature-based methods to segment rooms, treating \texttt{doors} and \texttt{windows} as part of the wall structure. For each pair of wall segments, we first compute their spatial overlap:

\begin{equation}
\begin{split}
c_1 &= \text{overlap}(W_{j-1}, W_j) \\
c_1 &\ge \tau_1
\end{split}
\label{eq:wall_overlap}
\end{equation}

where $W_{j-1},W_j$ is the wall segment output by neural network, $c_1$ denotes the overlap ratio between $W_{j-1}$ and $W_j$, and $\tau_1$ is a predefined threshold. Next, we evaluate their orientation similarity by computing the cosine similarity:

\begin{equation}
\begin{split}
c_2 &= \cos(V_{j-1}, V_j) \\
c_2 &\ge \tau_2
\end{split}
\label{eq:wall_angle}
\end{equation}
where $V_{j-1}$ and $V_j$ are the normal vector of wall segment.
If both conditions (\ref{eq:wall_overlap}) and (\ref{eq:wall_angle}) are satisfied, the two wall segments are merged.

For ceilings and floors, we check the vertical alignment of their center points. Let $Z_i$ and $Z_j$ denote the $z$-coordinates of their centroids:

\begin{equation}
\begin{split}
c_3 &= |Z_i - Z_j| \\
c_3 &\le \tau_3
\end{split}
\label{eq:vertical}
\end{equation}

If the height difference $c_3$ is below the threshold $\tau_3$, the surfaces are considered part of the same plane and are merged accordingly.
If a newly detected wall segment does not satisfy the above merging conditions with any existing room, it is treated as the boundary of a new room. The merging process is repeated iteratively until no new point cloud data is available. Its completeness is ensured by the long-range sensing capability of LiDAR and the iterative room segmentation algorithm, which together enable rapid extraction of room-level geometric structures. For each room, we compute the axis-aligned bounding box (AABB) of the merged wall segments by recording the minimum and maximum values along the $x$ and $y$ dimensions. This bounding box $B$ is dynamically maintained and later used to assist in object-instance construction and association within each room.

\vspace{-10pt}
\subsection{Instance-Level Semantic Extraction}
\label{SECTION III-B}

Instance-level semantic extraction is critical for spatial intelligence. While recent VFMs enable open-vocabulary understanding, their sensitivity to viewpoint, illumination, and occlusion still limits reliable segmentation in real-world settings.

\begin{figure}[ht]
    \centering
    \includegraphics[width=1\linewidth]{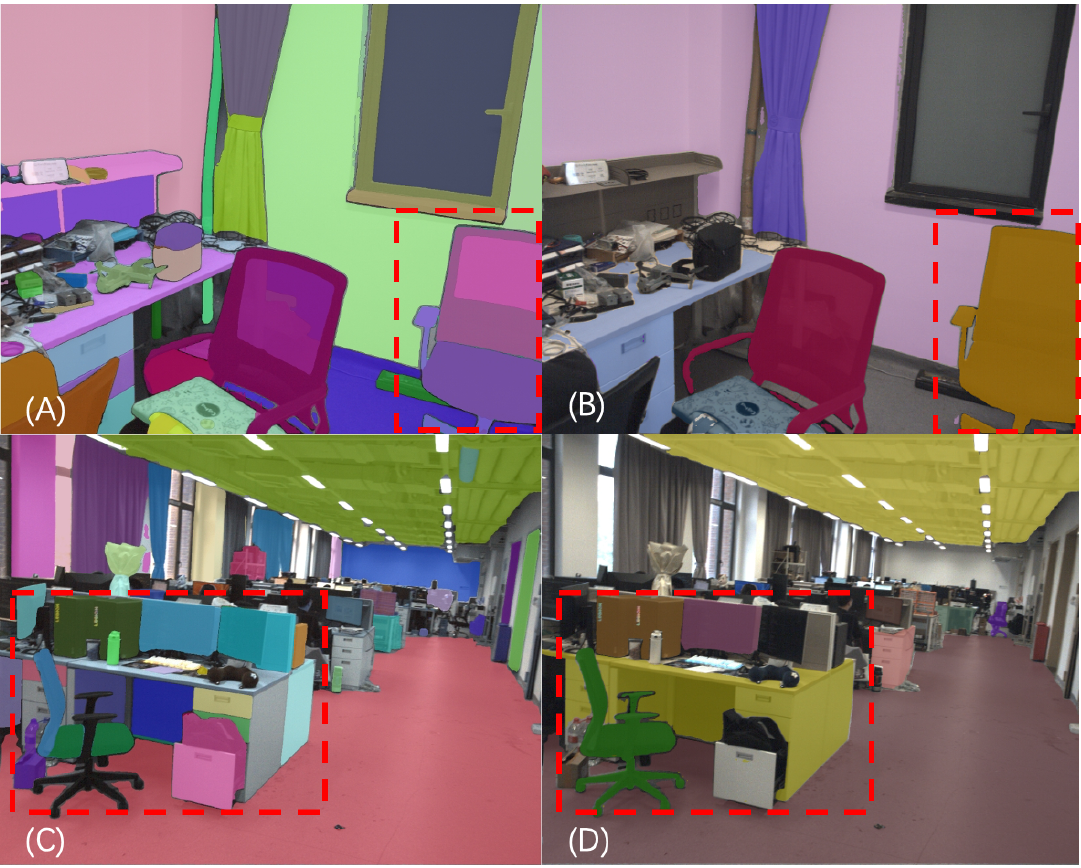}
    \caption{Comparison between SAM-only segmentation and our proposed method. (A) and (C) show results from direct SAM segmentation, while (B) and (D) are generated by our multi-stage pipeline. SAM tends to over-segment fine-grained structures, leading to semantic inconsistencies across viewpoints.}
    \label{fig:sam_compare}
\end{figure}

To address these issues, we adopt a multi-stage perception pipeline~\cite{ren2024grounded}. Initially, the RAM module interprets the scene linguistically by generating contextual prompts from the input RGB image. These prompts guide Grounding DINO to localize semantic objects via bounding boxes. Subsequently, SAM refines these regions to produce high-quality segmentation masks. Compared with repeated standalone SAM inferences, our composite pipeline demonstrates superior stability and accuracy, as illustrated in Fig.~\ref{fig:sam_compare}.

Following mask generation, we will assign the 2d mask with CLIP feature. We enhance semantic representation by extracting CLIP features from three complementary modalities: $\mathcal{L}_1$ from the masked RGB region, $\mathcal{L}_2$ from the binary mask, and $\mathcal{L}_3$ from the boundary-enhanced masked region. These features are linearly fused to yield a more robust and discriminative embedding:

\begin{equation}
\mathcal{L} = \alpha_1 \mathcal{L}_1 + \alpha_2 \mathcal{L}_2 + \alpha_3 \mathcal{L}_3
\label{eq:clip_fusion}
\end{equation}

where $\alpha_i$ ($i=1,2,3$) are predefined weights satisfying $\sum_{i=1}^{3} \alpha_i = 1$. This strategy improves retrieval fidelity under real-world conditions~\cite{werby2024hierarchical}.

To integrate semantic information into 3D space, we project each mask to a point cloud using the corresponding depth image. This step associates each 3D point with its semantic label, forming a semantically enriched point cloud. However, due to sensor noise and depth ambiguity, artifacts such as floating or misaligned points often arise. To suppress these, we apply DBSCAN clustering to filter out sparse outliers, improving spatial coherence and ensuring cleaner integration into downstream scene graphs.

\subsection{Instance Semantic Fusion}
\label{SECTION III-C}
After obtaining semantically labeled 3D point clouds, we aim to merge observations into consistent object instances. However, occlusions and viewpoint variations often cause fragmented segments, making accurate association difficult. Existing methods typically rely on geometric overlap with exhaustive pairwise comparisons, resulting in high computational overhead and limiting scalability.

In our method, instance fusion is guided by both semantic and geometric cues within spatially constrained regions. Building on the room segmentation described in Section~\ref{SECTION III-A}, we pre-classify all reconstructed point clouds according to their associated rooms. We assume that reliable instance observations originate from within the same room—a simplification that, while potentially omitting a small number of cross-room observations, has minimal impact (see Experiment~\ref{SECTION IV-A} for analysis).
This room-based partitioning enables parallel instance fusion across rooms, significantly improving computational efficiency and scalability. 
Furthermore, instead of relying on conventional sequential mask fusion, we adopt a mask-query-based fusion strategy as proposed in~\cite{yang2023sam3dsegment3dscenes}. This approach allows new observations to be matched against a compact memory of existing instances, avoiding exhaustive pairwise comparisons and improving fusion efficiency.

To ensure the accuracy of instance fusion, we combine both geometric and semantic cues. Traditional fusion methods typically rely solely on point cloud overlap, which can lead to incorrect merges~(\textit{e.g.}, books on a desk being fused with the desk itself). To address this, we introduce a dual-criteria fusion strategy. Given two candidate 3d mask point cloud $M_{j-1},M_{j}$, we compute their geometric overlap score
\begin{equation}
\begin{split}
i_1 &= \text{overlap}(M_{j-1}, M_j) \\
i_2 &\ge \tau_g
\end{split}
\label{eq:iou}
\end{equation}
where $i_1$ must satisfy $i_1 \ge \tau_g$, with $\tau_g$ denoting the geometric threshold.
Simultaneously, we evaluate the semantic similarity between their associated features using cosine similarity:
\begin{equation}
\begin{split}
i_2 &= \text{consine}(M_{j-1}, M_j) \\
i_2 &\ge \tau_s
\end{split}
\label{eq:simlarity}
\end{equation}
and require $i_2 \ge \tau_s$, where $\tau_s$ is the semantic threshold. Fusion is performed only when both criteria are satisfied. In our implementation, the overlap threshold $\tau_g$ is set to 0.3, while the semantic similarity threshold $\tau_s$ is set to 0.8. This approach ensures that even small objects with precise semantics can be preserved, avoiding over-aggregation.

\subsection{Scene Graph Construction}
\label{SECTION III-D}
After constructing the room and instance-level semantics, we proceed to build the scene graph. Our scene graph is organized into three hierarchical levels: building, room, and instance. First, the room segmentation process yields a set of rooms, ${R_1, \ldots, R_m}$. For determining the semantic type of each room, directly averaging the CLIP features of contained objects often leads to bias due to the presence of common small objects. To address this, we weight each instance’s CLIP feature by its volume, giving larger objects greater influence as they tend to be more representative. The resulting aggregated feature is then compared against a predefined set of room category prototypes~\texttt{Kitchen,Office,Dining room,Bedroom,Bathroom}, and the label with the highest similarity is assigned to the room.

At the instance level, we assign objects to specific rooms based on their bounding boxes, establishing a belongs-to relationship between objects and rooms. Once these steps are completed, a comprehensive 3D scene graph is obtained.

This scene graph not only provides a rich semantic understanding of the environment but also aligns seamlessly with natural human language, enabling more effective human-robot interaction. Furthermore, it supports direct querying via large language models (see Experiment~\ref{SECTION IV-C}), opening up limitless possibilities for future applications.

\begin{figure*}
    \centering
    \includegraphics[width=1\linewidth]{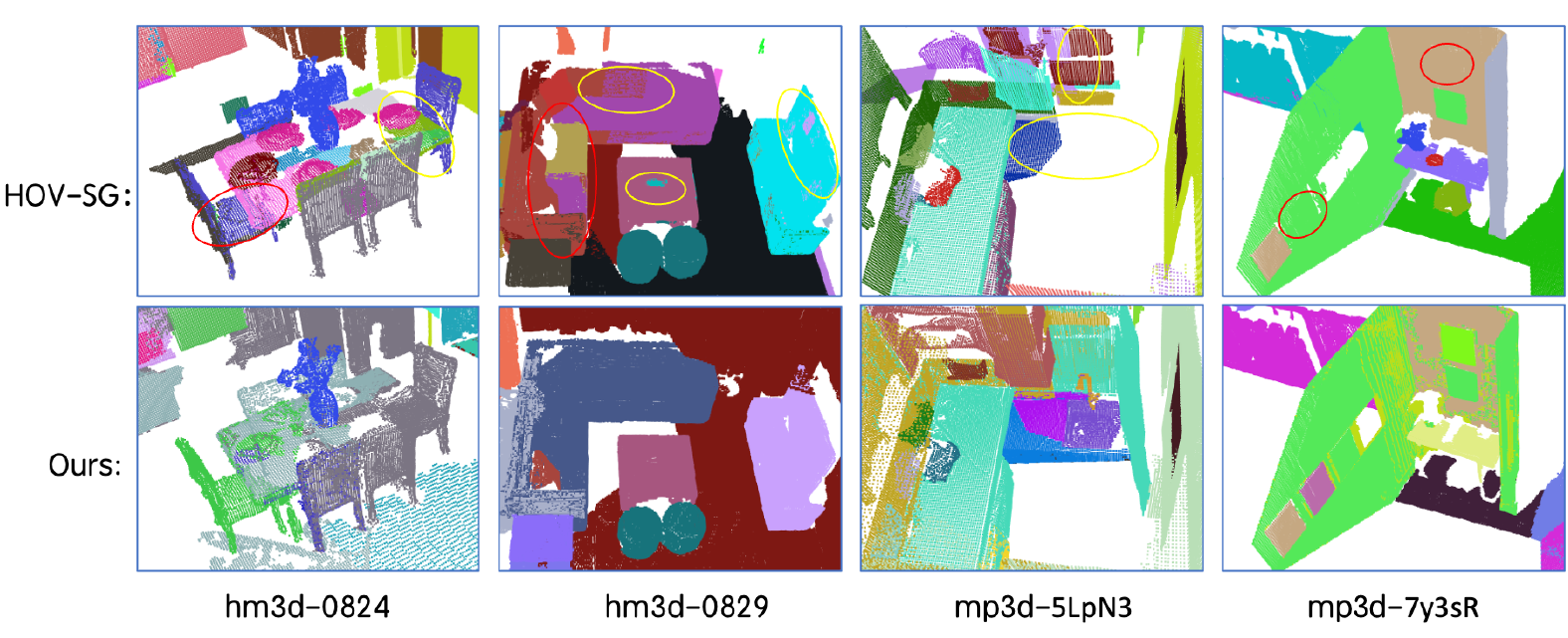}
    \caption{Result of scene construction. The first row is rebuilt by HOV-SG~\cite{werby2024hierarchical}, the second row is ours. The falsely predicted semantic classes in (a) and (b) are highlighted in red circles, while spatial conflicted semantics are in yellow. All instances are colored randomly.} 
    \label{fig: simulation result}
\end{figure*}

\section{Experiment Verification}

\label{SECTION IV}
To evaluate the efficiency and robustness of our system, we conduct a series of experiments across both simulated and real-world settings: Experiment 1: scene graph construction in simulation.
We evaluate the efficiency and scalability of our method in a controlled simulation environment, demonstrating its ability to build accurate semantic scene graphs in large-scale spaces~\ref{SECTION IV-A}. Real-world scene reconstruction will be shown in Experiment 2~\ref{SECTION IV-B}. We deploy our system in real indoor environments to verify its robustness, adaptability, and reliability under practical conditions.
In Experiment 3~\ref{SECTION IV-C}: Semantic navigation using scene graphs. We leverage the constructed scene graph to guide a robot via natural language commands, validating the effectiveness of our proposed vision-language alignment in downstream navigation tasks.

\subsection{Simulation-based Reconstruction Evaluation}
\label{SECTION IV-A}
In this section, we evaluate the performance of the proposed method in simulated environments, primarily focusing on the reconstruction accuracy and effectiveness. 

We conducted both comparative and ablation studies on the HM3D and MP3D datasets. Our approach was benchmarked against the state-of-the-art HOV-SG~\cite{werby2024hierarchical} framework, with additional evaluation performed to assess the impact of removing room-level priors.
The evaluation centers on three core metrics: reconstruction speed, semantic accurancy in top5 prediction, and average precision for class-agnostic instance segmentation. Representative results from the simulation experiments are shown in Fig.~\ref{fig: simulation result}. All experiments were conducted using an NVIDIA RTX 4090 GPU and an Intel Xeon Platinum 8370C CPU @ 2.80GHz.

Surprisingly, despite our method discarding cross-room observations during instance fusion, the overall semantic accuracy remains on par with HOV-SG. In certain scenes, our approach even outperforms state-of-the-art significantly. We attribute this improvement to the fact that HOV-SG relies heavily on geometric heuristics for merging instances, often neglecting valuable semantic priors, whereas our method leverages both.

Furthermore, our approach achieves an order-of-magnitude speedup in reconstruction time. In our ablation study, we observed that introducing room-level segmentation and enabling parallel instance fusion yields a 78.2\% speed gain compared to a naive serial merging pipeline. These results clearly demonstrate the efficiency and robustness of our system. Detailed comparisons are presented in Table~\ref{experiment 1}.

\begin{table}[h]
\centering
\caption{\textbf{Simulation Reconstruction Result}}\label{experiment 1}
\resizebox{\columnwidth}{!}{
\begin{tabular}{lcccccc}
\toprule & \multicolumn{3}{c}{HM3DSem scene 00824} & \multicolumn{3}{c}{HM3DSem scene 00829}   \\
\cmidrule(r){2-4} \cmidrule(r){5-7}   
& \textbf{Time}[s] & $\textbf{top}_\textbf{5}$ & \textbf{AP} &  \textbf{Time}[s] & $\textbf{top}_\textbf{5}$  & \textbf{AP} \\
\midrule
IRS (ours)  & \textbf{118} & 25.53 & \textbf{28.03} & \textbf{80} & 24.19 & \textbf{31.11} \\
IRS w/o room & 517 & \textbf{28.13} & 26.34 & 515 & \textbf{25.37} & 30.04 \\
HOV-SG & 5221 & 18.43 & 25.12 & 4263 & 21.05 & 24.62\\
\bottomrule
\end{tabular}
}

\end{table}

\subsection{Real-World Performance Evaluation}
\label{SECTION IV-B}

\begin{figure}
    \centering
    \includegraphics[width=1\linewidth]{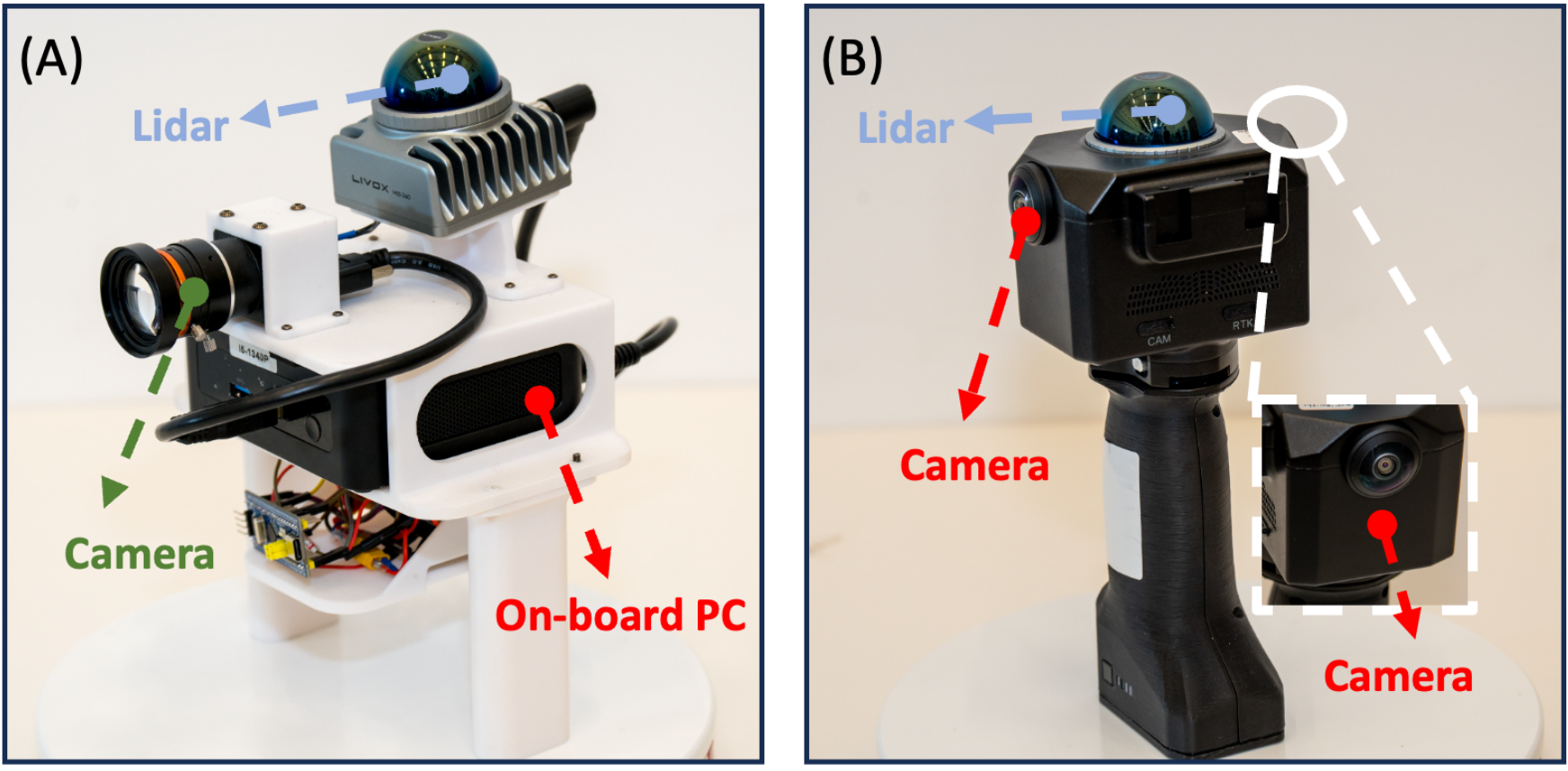}
    \caption{  Data collection setups used in our real-world experiments. 
  (A) shows our custom-built sensor, which integrates a LiDAR and a monocular camera.
  (B) displays a commercial handheld device equipped with a LiDAR and two fisheye cameras.}
    \label{fig: equipment}
\end{figure}

This section evaluates the real-world performance of our proposed system, focusing on its robustness and accuracy in constructing scene graphs under practical conditions. To thoroughly validate the effectiveness of our approach, we conducted experiments using two distinct data acquisition setups: a commercial handheld device\footnote{\url{https://www.manifold.com.co/pocket}} and a custom-built sensing platform (see Fig.~\ref{fig: equipment}). Both setups were deployed in the same real-world environment—our own office, which is intentionally unstructured and irregular, differing from standard benchmark environments. We used FAST-LIVO2~\cite{zheng2024fast} to collect synchronized LiDAR point clouds and rgb data. Scene graphs were then constructed based on the collected data.
We leveraged GPT-4~\cite{achiam2023gpt} to parse natural language queries. GPT extracts structured elements such as building names, room types, and object descriptions from the input query, encodes them into CLIP embeddings, and retrieves relevant entities in the scene graph based on similarity matching~(any large language model is theoretically feasible).

Additionally, to verify the generalization capabilities of our system beyond common object categories, we placed uncommon objects (\textit{e.g.}, drones) in the scene. These items, which are rarely labeled in standard datasets, were successfully identified and retrieved via open-vocabulary queries—further validating the semantic richness and flexibility of our system.
\begin{figure}[ht]
    \centering
    \includegraphics[width=1\linewidth]{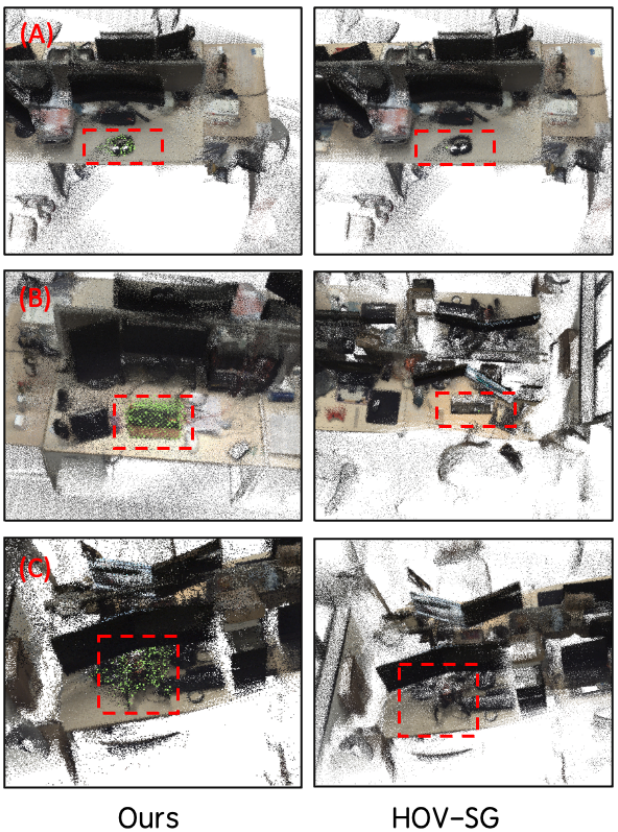}
    \caption{Real-world semantic query results. Green highlights indicate successful retrievals, with a greater presence of green reflecting higher semantic consistency. Our method (left) is compared with HOV-SG (right) across three queries: (A) “find headphone”, (B) “find the keyboard”, and (C) “find the aerial drone”. Under cluttered and complex conditions, our approach yields more consistent semantic matches and cleaner instance boundaries.
    }
    \label{fig:real world construction}
    \vspace{-10pt}
\end{figure}
Figure~\ref{fig:real world construction} presents the scene graph results constructed in a real-world environment. The right side shows the output of HOV-SG, while the left side displays the result of our proposed method. It is evident that HOV-SG suffers from severe semantic fragmentation issues and yields low accuracy during semantic querying.

We attribute this problem primarily to the complexity of the testing environment: the desktop is cluttered and disorganized, with significant occlusions and variations in object placement. Under such conditions, SAM fails to maintain consistent segmentation across consecutive frames. This inconsistency propagates through the instance fusion stage and CLIP-based feature alignment, ultimately preventing the construction of complete and coherent object instances, thereby degrading the overall quality of the scene graph.

In contrast, our method addresses this issue at the source. By improving the initial quality and consistency of the segmentation masks, we ensure more accurate fusion and robust semantic alignment. Experimental results demonstrate that our approach produces more reliable masks and successfully builds more accurate and stable 3D scene graphs.

\subsection{Robotic Semantic Navigation}
\label{SECTION IV-C}
To further demonstrate the semantic reasoning capabilities and generalization of our system, we conducted an additional experiment focused on cross-room navigation. This task tests the robot’s ability to interpret high-level human language commands and navigate through multiple rooms based on semantic understanding.

Building on the setup in Experiment~\ref{SECTION IV-B}, we constructed a scene graph using data collected in a real-world, unstructured office-laboratory environment. The constructed scene graph is illustrated in Fig.~\ref{fig:1}. In this scenario, the robot was initially positioned in the office, and the following command was issued: \textit{“Go to the TV in the laboratory”}. This instruction requires the robot to reason over both semantic and spatial context, identifying the correct object and planning a trajectory across rooms.

\begin{figure}
    \centering
    \includegraphics[width=1\linewidth]{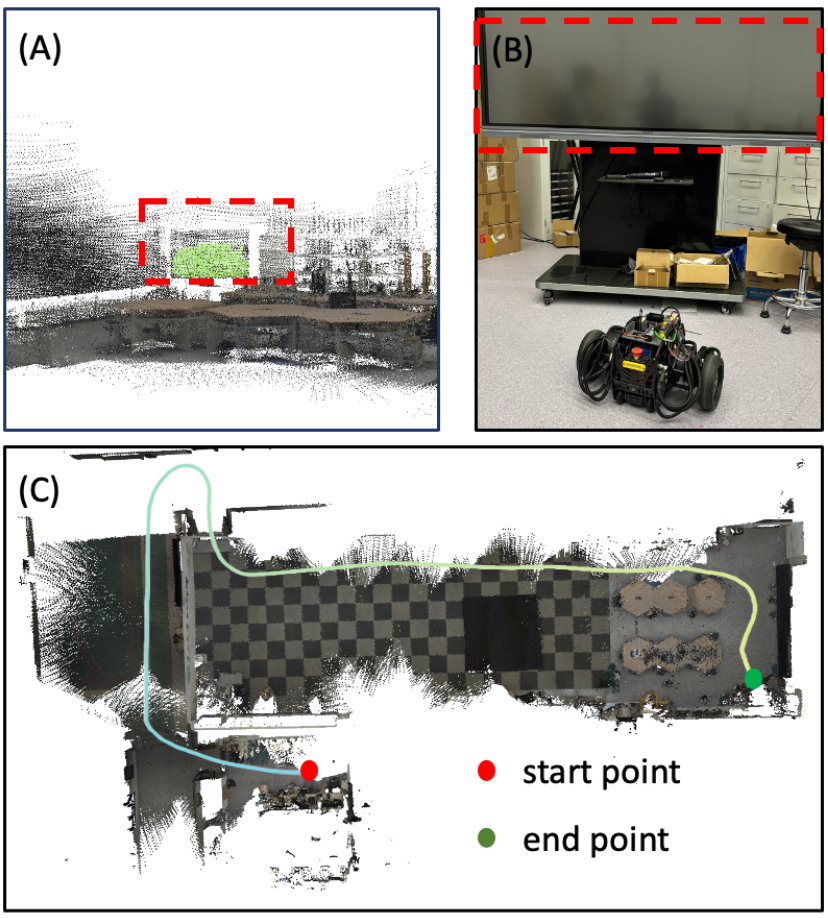}
    \caption{Real-world cross-room semantic navigation experiment. The robot is initially located in the office and receives the command “Go to the TV in the laboratory.”}
    \label{fig: navigation}
    \vspace{-10pt}
\end{figure}

To evaluate task success, we defined completion as reaching within 0.5 meter of the target object. As shown in Fig.~\ref{fig: navigation}, our system successfully parsed the natural language command, extracted semantic cues using an LLM, identified the relevant room and object in the scene graph via CLIP-based similarity matching, and navigated autonomously to the specified location.

Unlike traditional approaches that rely solely on metric coordinates, our system leverages the constructed scene graph to enable semantic-level understanding and flexible navigation. This experiment highlights the robustness and potential of our method in realistic and open-vocabulary navigation scenarios that require global scene comprehension.

\section{CONCLUSION AND FUTURE WORK}

 This paper presents the IRS system architecture, which leverages LiDAR to rapidly acquire room-level semantic information, serving as the foundation for instance construction. It then uses multi-level VFMs to obtain semantic data from RGB inputs and builds a 3d scene graph. Compared to state-of-the-art scene graph construction methods, our approach significantly improves both effectiveness and efficiency without compromising accuracy. Additionally, we have demonstrated its potential for navigation, where it can reach targets using human language commands without requiring precise pose information. In the future, we aim to refine this framework, further enhancing its construction speed and extending its applications in semantic navigation. We also plan to explore end-to-end semantic navigation without relying on map-based representations.

\label{SECTION V}

\bibliography{RAL2025_reference} 

\begin{thebibliography}{10}
\providecommand{\url}[1]{#1}
\csname url@rmstyle\endcsname
\providecommand{\newblock}{\relax}
\providecommand{\bibinfo}[2]{#2}
\providecommand\BIBentrySTDinterwordspacing{\spaceskip=0pt\relax}
\providecommand\BIBentryALTinterwordstretchfactor{4}
\providecommand\BIBentryALTinterwordspacing{\spaceskip=\fontdimen2\font plus
\BIBentryALTinterwordstretchfactor\fontdimen3\font minus \fontdimen4\font\relax}
\providecommand\BIBforeignlanguage[2]{{%
\expandafter\ifx\csname l@#1\endcsname\relax
\typeout{** WARNING: IEEEtran.bst: No hyphenation pattern has been}%
\typeout{** loaded for the language `#1'. Using the pattern for}%
\typeout{** the default language instead.}%
\else
\language=\csname l@#1\endcsname
\fi
#2}}

\bibitem{deng2025whole}
W.~Deng, H.~Chen, B.~Ye, H.~Chen, and X.~Lyu, ``Whole-body integrated motion planning for aerial manipulators,'' \emph{arXiv preprint arXiv:2501.06493}, 2025.

\bibitem{han2023efficient}
Z.~Han, Y.~Wu, T.~Li, L.~Zhang, L.~Pei, L.~Xu, C.~Li, C.~Ma, C.~Xu, S.~Shen, \emph{et~al.}, ``An efficient spatial-temporal trajectory planner for autonomous vehicles in unstructured environments,'' \emph{IEEE Transactions on Intelligent Transportation Systems}, vol.~25, no.~2, pp. 1797--1814, 2023.

\bibitem{gervet2023navigating}
T.~Gervet, S.~Chintala, D.~Batra, J.~Malik, and D.~S. Chaplot, ``Navigating to objects in the real world,'' \emph{Science Robotics}, vol.~8, no.~79, p. eadf6991, 2023.

\bibitem{wu2021scenegraphfusion}
S.-C. Wu, J.~Wald, K.~Tateno, N.~Navab, and F.~Tombari, ``Scenegraphfusion: Incremental 3d scene graph prediction from rgb-d sequences,'' in \emph{Proceedings of the IEEE/CVF Conference on Computer Vision and Pattern Recognition}, 2021, pp. 7515--7525.

\bibitem{greve2024collaborative}
E.~Greve, M.~B{\"u}chner, N.~V{\"o}disch, W.~Burgard, and A.~Valada, ``Collaborative dynamic 3d scene graphs for automated driving,'' in \emph{2024 IEEE International Conference on Robotics and Automation (ICRA)}.\hskip 1em plus 0.5em minus 0.4em\relax IEEE, 2024, pp. 11\,118--11\,124.

\bibitem{yang2023sam3dsegment3dscenes}
\BIBentryALTinterwordspacing
Y.~Yang, X.~Wu, T.~He, H.~Zhao, and X.~Liu, ``Sam3d: Segment anything in 3d scenes,'' 2023. [Online]. Available: \url{https://arxiv.org/abs/2306.03908}
\BIBentrySTDinterwordspacing

\bibitem{rosinol2020kimera}
A.~Rosinol, M.~Abate, Y.~Chang, and L.~Carlone, ``Kimera: an open-source library for real-time metric-semantic localization and mapping,'' in \emph{2020 IEEE International Conference on Robotics and Automation (ICRA)}.\hskip 1em plus 0.5em minus 0.4em\relax IEEE, 2020, pp. 1689--1696.

\bibitem{mccormac2018fusion++}
J.~McCormac, R.~Clark, M.~Bloesch, A.~Davison, and S.~Leutenegger, ``Fusion++: Volumetric object-level slam,'' in \emph{2018 international conference on 3D vision (3DV)}.\hskip 1em plus 0.5em minus 0.4em\relax IEEE, 2018, pp. 32--41.

\bibitem{mccormac2017semanticfusion}
J.~McCormac, A.~Handa, A.~Davison, and S.~Leutenegger, ``Semanticfusion: Dense 3d semantic mapping with convolutional neural networks,'' in \emph{2017 IEEE International Conference on Robotics and automation (ICRA)}.\hskip 1em plus 0.5em minus 0.4em\relax IEEE, 2017, pp. 4628--4635.

\bibitem{kirillov2023segany}
A.~Kirillov, E.~Mintun, N.~Ravi, H.~Mao, C.~Rolland, L.~Gustafson, T.~Xiao, S.~Whitehead, A.~C. Berg, W.-Y. Lo, P.~Doll{\'a}r, and R.~Girshick, ``Segment anything,'' \emph{arXiv:2304.02643}, 2023.

\bibitem{liu2023grounding}
S.~Liu, Z.~Zeng, T.~Ren, F.~Li, H.~Zhang, J.~Yang, C.~Li, J.~Yang, H.~Su, J.~Zhu, \emph{et~al.}, ``Grounding dino: Marrying dino with grounded pre-training for open-set object detection,'' \emph{arXiv preprint arXiv:2303.05499}, 2023.

\bibitem{werby2024hierarchical}
A.~Werby, C.~Huang, M.~B{\"u}chner, A.~Valada, and W.~Burgard, ``Hierarchical open-vocabulary 3d scene graphs for language-grounded robot navigation,'' in \emph{First Workshop on Vision-Language Models for Navigation and Manipulation at ICRA 2024}, 2024.

\bibitem{rosinol2021kimera}
A.~Rosinol, A.~Violette, M.~Abate, N.~Hughes, Y.~Chang, J.~Shi, A.~Gupta, and L.~Carlone, ``Kimera: From slam to spatial perception with 3d dynamic scene graphs,'' \emph{The International Journal of Robotics Research}, vol.~40, no. 12-14, pp. 1510--1546, 2021.

\bibitem{chang2023goatthing}
\BIBentryALTinterwordspacing
M.~Chang, T.~Gervet, M.~Khanna, S.~Yenamandra, D.~Shah, S.~Y. Min, K.~Shah, C.~Paxton, S.~Gupta, D.~Batra, R.~Mottaghi, J.~Malik, and D.~S. Chaplot, ``Goat: Go to any thing,'' 2023. [Online]. Available: \url{https://arxiv.org/abs/2311.06430}
\BIBentrySTDinterwordspacing

\bibitem{chaplot2020learningexploreusingactive}
\BIBentryALTinterwordspacing
D.~S. Chaplot, D.~Gandhi, S.~Gupta, A.~Gupta, and R.~Salakhutdinov, ``Learning to explore using active neural slam,'' 2020. [Online]. Available: \url{https://arxiv.org/abs/2004.05155}
\BIBentrySTDinterwordspacing

\bibitem{lu2024semantics}
L.~Lu, Y.~Zhang, P.~Zhou, J.~Qi, Y.~Pan, C.~Fu, and J.~Pan, ``Semantics-aware receding horizon planner for object-centric active mapping,'' \emph{IEEE Robotics and Automation Letters}, 2024.

\bibitem{whelan2016elasticfusion}
T.~Whelan, R.~F. Salas-Moreno, B.~Glocker, A.~J. Davison, and S.~Leutenegger, ``Elasticfusion: Real-time dense slam and light source estimation,'' \emph{The International Journal of Robotics Research}, vol.~35, no.~14, pp. 1697--1716, 2016.

\bibitem{liu2024fm}
C.~Liu, K.~Wang, J.~Shi, Z.~Qiao, and S.~Shen, ``Fm-fusion: Instance-aware semantic mapping boosted by vision-language foundation models,'' \emph{IEEE Robotics and Automation Letters}, vol.~9, no.~3, pp. 2232--2239, 2024.

\bibitem{yang2018graph}
J.~Yang, J.~Lu, S.~Lee, D.~Batra, and D.~Parikh, ``Graph r-cnn for scene graph generation,'' in \emph{Proceedings of the European conference on computer vision (ECCV)}, 2018, pp. 670--685.

\bibitem{radford2021learningtransferablevisualmodels}
\BIBentryALTinterwordspacing
A.~Radford, J.~W. Kim, C.~Hallacy, A.~Ramesh, G.~Goh, S.~Agarwal, G.~Sastry, A.~Askell, P.~Mishkin, J.~Clark, G.~Krueger, and I.~Sutskever, ``Learning transferable visual models from natural language supervision,'' 2021. [Online]. Available: \url{https://arxiv.org/abs/2103.00020}
\BIBentrySTDinterwordspacing

\bibitem{yokoyama2024vlfm}
N.~Yokoyama, S.~Ha, D.~Batra, J.~Wang, and B.~Bucher, ``Vlfm: Vision-language frontier maps for zero-shot semantic navigation,'' in \emph{2024 IEEE International Conference on Robotics and Automation (ICRA)}.\hskip 1em plus 0.5em minus 0.4em\relax IEEE, 2024, pp. 42--48.

\bibitem{yin2024sgnavonline3dscene}
\BIBentryALTinterwordspacing
H.~Yin, X.~Xu, Z.~Wu, J.~Zhou, and J.~Lu, ``Sg-nav: Online 3d scene graph prompting for llm-based zero-shot object navigation,'' 2024. [Online]. Available: \url{https://arxiv.org/abs/2410.08189}
\BIBentrySTDinterwordspacing

\bibitem{armeni20193d}
I.~Armeni, Z.-Y. He, J.~Gwak, A.~R. Zamir, M.~Fischer, J.~Malik, and S.~Savarese, ``3d scene graph: A structure for unified semantics, 3d space, and camera,'' in \emph{Proceedings of the IEEE/CVF international conference on computer vision}, 2019, pp. 5664--5673.

\bibitem{liu2024slideslam}
X.~Liu, J.~Lei, A.~Prabhu, Y.~Tao, I.~Spasojevic, P.~Chaudhari, N.~Atanasov, and V.~Kumar, ``Slideslam: Sparse, lightweight, decentralized metric-semantic slam for multi-robot navigation,'' \emph{arXiv preprint arXiv:2406.17249}, 2024.

\bibitem{hughes2022hydra}
N.~Hughes, Y.~Chang, and L.~Carlone, ``Hydra: A real-time spatial perception system for 3d scene graph construction and optimization,'' \emph{arXiv preprint arXiv:2201.13360}, 2022.

\bibitem{bormann2016room}
R.~Bormann, F.~Jordan, W.~Li, J.~Hampp, and M.~H{\"a}gele, ``Room segmentation: Survey, implementation, and analysis,'' in \emph{2016 IEEE international conference on robotics and automation (ICRA)}.\hskip 1em plus 0.5em minus 0.4em\relax IEEE, 2016, pp. 1019--1026.

\bibitem{luperto2019predicting}
M.~Luperto, V.~Arcerito, and F.~Amigoni, ``Predicting the layout of partially observed rooms from grid maps,'' in \emph{2019 International Conference on Robotics and Automation (ICRA)}.\hskip 1em plus 0.5em minus 0.4em\relax IEEE, 2019, pp. 6898--6904.

\bibitem{luperto2022robust}
M.~Luperto, T.~P. Kucner, A.~Tassi, M.~Magnusson, and F.~Amigoni, ``Robust structure identification and room segmentation of cluttered indoor environments from occupancy grid maps,'' \emph{IEEE Robotics and Automation Letters}, vol.~7, no.~3, pp. 7974--7981, 2022.

\bibitem{he2021hierarchical}
Z.~He, H.~Sun, J.~Hou, Y.~Ha, and S.~Schwertfeger, ``Hierarchical topometric representation of 3d robotic maps,'' \emph{Autonomous Robots}, vol.~45, no.~5, pp. 755--771, 2021.

\bibitem{ambrucs2017automatic}
R.~Ambru{\c{s}}, S.~Claici, and A.~Wendt, ``Automatic room segmentation from unstructured 3-d data of indoor environments,'' \emph{IEEE Robotics and Automation Letters}, vol.~2, no.~2, pp. 749--756, 2017.

\bibitem{blochliger2018topomap}
F.~Blochliger, M.~Fehr, M.~Dymczyk, T.~Schneider, and R.~Siegwart, ``Topomap: Topological mapping and navigation based on visual slam maps,'' in \emph{2018 IEEE International Conference on Robotics and Automation (ICRA)}.\hskip 1em plus 0.5em minus 0.4em\relax IEEE, 2018, pp. 3818--3825.

\bibitem{chen2022fast}
X.~Chen, B.~Zhou, J.~Lin, Y.~Zhang, F.~Zhang, and S.~Shen, ``Fast 3d sparse topological skeleton graph generation for mobile robot global planning,'' in \emph{2022 IEEE/RSJ International Conference on Intelligent Robots and Systems (IROS)}.\hskip 1em plus 0.5em minus 0.4em\relax IEEE, 2022, pp. 10\,283--10\,289.

\bibitem{qi2017pointnet}
C.~R. Qi, H.~Su, K.~Mo, and L.~J. Guibas, ``Pointnet: Deep learning on point sets for 3d classification and segmentation,'' in \emph{Proceedings of the IEEE conference on computer vision and pattern recognition}, 2017, pp. 652--660.

\bibitem{zhang2019review}
J.~Zhang, X.~Zhao, Z.~Chen, and Z.~Lu, ``A review of deep learning-based semantic segmentation for point cloud,'' \emph{IEEE access}, vol.~7, pp. 179\,118--179\,133, 2019.

\bibitem{kolodiazhnyi2024oneformer3d}
M.~Kolodiazhnyi, A.~Vorontsova, A.~Konushin, and D.~Rukhovich, ``Oneformer3d: One transformer for unified point cloud segmentation,'' in \emph{Proceedings of the IEEE/CVF Conference on Computer Vision and Pattern Recognition}, 2024, pp. 20\,943--20\,953.

\bibitem{tversky2003structures}
B.~Tversky, ``Structures of mental spaces: How people think about space,'' \emph{Environment and behavior}, vol.~35, no.~1, pp. 66--80, 2003.

\bibitem{ren2024grounded}
T.~Ren, S.~Liu, A.~Zeng, J.~Lin, K.~Li, H.~Cao, J.~Chen, X.~Huang, Y.~Chen, F.~Yan, \emph{et~al.}, ``Grounded sam: Assembling open-world models for diverse visual tasks,'' \emph{arXiv preprint arXiv:2401.14159}, 2024.

\bibitem{zheng2024fast}
C.~Zheng, W.~Xu, Z.~Zou, T.~Hua, C.~Yuan, D.~He, B.~Zhou, Z.~Liu, J.~Lin, F.~Zhu, \emph{et~al.}, ``Fast-livo2: Fast, direct lidar-inertial-visual odometry,'' \emph{IEEE Transactions on Robotics}, 2024.

\bibitem{achiam2023gpt}
J.~Achiam, S.~Adler, S.~Agarwal, L.~Ahmad, I.~Akkaya, F.~L. Aleman, D.~Almeida, J.~Altenschmidt, S.~Altman, S.~Anadkat, \emph{et~al.}, ``Gpt-4 technical report,'' \emph{arXiv preprint arXiv:2303.08774}, 2023.

\end{thebibliography}

\end{document}